%% file: main.tex
\documentclass[conference]{IEEEtran}
\IEEEoverridecommandlockouts
\usepackage{cite}
\usepackage{amsmath,amssymb,amsfonts}
\usepackage{algorithmic}
\usepackage{graphicx}
\usepackage{textcomp}
\usepackage{xcolor}
\usepackage{algorithm}
\usepackage{algorithmic}
\usepackage{rotating}
\usepackage{booktabs}
\usepackage{multirow}
\usepackage{bbm}

\makeatletter
\DeclareFontShape{OT1}{ptm}{m}{scit}{<->ssub*ptm/m/sc}{ }
\DeclareFontShape{OT1}{ptm}{b}{scit}{<->ssub*ptm/b/sc}{ }
\makeatother

\def\BibTeX{{\rm B\kern-.05em{\sc i\kern-.025em b}\kern-.08em
    T\kern-.1667em\lower.7ex\hbox{E}\kern-.125emX}}
\begin{document}

\newcommand{\bea}{\begin{eqnarray}} 
\newcommand{\eea}{\end{eqnarray}}
\newcommand{\e}[1]{\times 10^{#1}}  
\newcommand{\loss}{\mathcal{L}}


\linespread{0.9}

\title{Efficient KernelSHAP Explanations for Patch-based 3D Medical Image Segmentation}

\author{
\IEEEauthorblockN{Ricardo Coimbra Brioso\IEEEauthorrefmark{1}, 
Giulio Sichili\IEEEauthorrefmark{1}, 
Damiano Dei\IEEEauthorrefmark{3}, 
Nicola Lambri\IEEEauthorrefmark{3}\IEEEauthorrefmark{4},\\ 
Pietro Mancosu\IEEEauthorrefmark{2}\IEEEauthorrefmark{3}, 
Marta Scorsetti\IEEEauthorrefmark{2}\IEEEauthorrefmark{3}, 
and Daniele Loiacono\IEEEauthorrefmark{1}}
\smallskip

\IEEEauthorblockA{\IEEEauthorrefmark{1}Dipartimento di Elettronica, Informazione e Bioingegneria, Politecnico di Milano, Milan, Italy}

\IEEEauthorblockA{\IEEEauthorrefmark{2}Department of Biomedical Sciences, Humanitas University, Pieve Emanuele, Milan, Italy}

\IEEEauthorblockA{\IEEEauthorrefmark{3}Radiotherapy and Radiosurgery Department, IRCCS Humanitas Research Hospital, Rozzano, Milan, Italy}

\IEEEauthorblockA{\IEEEauthorrefmark{4}Scuola di Specializzazione di Fisica Medica, Università Degli Studi di Milano, Milan, Italy}

\IEEEauthorblockA{Email: ricardo.brioso@polimi.it, daniele.loiacono@polimi.it}
}

\maketitle

\begin{abstract}
Perturbation-based explainability methods such as KernelSHAP provide model-agnostic attributions but are typically impractical for patch-based 3D medical image segmentation due to the large number of coalition evaluations and the high cost of sliding-window inference. We present an efficient KernelSHAP framework for volumetric CT segmentation that restricts computation to a user-defined region of interest and its receptive-field support, and accelerates inference via patch logit caching, reusing baseline predictions for unaffected patches while preserving nnU-Net’s fusion scheme. To enable clinically meaningful attributions, we compare three automatically generated feature abstractions within the receptive-field crop: whole-organ units, regular FCC supervoxels, and hybrid organ-aware supervoxels, and we study multiple aggregation/value functions targeting stabilizing evidence (TP/Dice/Soft Dice) or false-positive behavior. Experiments on whole-body CT segmentations show that caching substantially reduces redundant computation (with computational savings ranging from 15\% to 30\%) and that faithfulness and interpretability exhibit clear trade-offs: regular supervoxels often maximize perturbation-based metrics but lack anatomical alignment, whereas organ-aware units yield more clinically interpretable explanations and are particularly effective for highlighting false-positive drivers under normalized metrics.
\end{abstract}


\section{Introduction}
\label{sec:introduction}
\input{sections/sec_introduction2.tex}

\section{Related Works}
\label{sec:related}
\input{sections/sec_related2.tex}

\section{Adapting KernelSHAP for Efficient 3D Segmentation Explainability}
\label{sec:methods}
\input{sections/sec_methods2.tex}


\section{Experimental Design}
\label{sec:exp_design_xai}
\input{sections/sec_design2.tex}

\section{Results}
\label{sec:results_xai}
\input{sections/sec_results2.tex}

\section{Conclusions} 
\label{sec:conclusion_xai}
\input{sections/sec_conclusions2.tex}

\section*{Acknowledgment}
The authors disclose the use of OpenAI ChatGPT (version 5.2) to assist with English-language editing and linguistic refinement of the manuscript. The use of AI was limited to language polishing, while all technical and scientific content was produced and verified by the authors.

\bibliographystyle{ieeetr} 
\bibliography{myrefs.bib}

\end{document}

%% file: sections/sec_introduction2.tex
Deep learning models have become the de-facto standard for medical image segmentation, enabling robust delineation of anatomical structures and pathological targets across modalities and clinical workflows. 
In radiotherapy treatment planning, in particular, accurate segmentation of organs-at-risk and target volumes directly impacts dose optimization and safety. 
Despite their strong empirical performance, deep learning segmentation models remain difficult to audit: errors can be subtle, spatially localized, and strongly influenced by contextual anatomy, acquisition artifacts, or model's biases. 
This motivates explainable AI (XAI) methods capable of providing faithful, clinically meaningful explanations of \emph{why} a 3D segmentation model produced a given mask rather than another.

Compared to classification, explainability for dense prediction is substantially less consolidated. Recent surveys highlight open challenges specific to segmentation, including (i) how to define interpretable regions of analysis in volumetric data, (ii) how to aggregate voxel-level evidence into robust scores, and (iii) how to preserve anatomical structure in explanations \cite{gipiskis_explainable_2024}. Gradient-based methods (e.g., segmentation-adapted Grad-CAM variants) are computationally efficient but may be layer-dependent and hard to interpret in multi-structure settings \cite{vinogradova_towards_2020,hasany_guided_2024,silva_fm-g-cam_2024}. 
Perturbation-based approaches, on the other hand, offer model-agnostic explanations and can be aligned with clinically meaningful units, but they are often prohibitively expensive in 3D---especially for patch-based inference pipelines such as nnU-Net \cite{isensee_nnu-net_2021}. In particular, Shapley-value methods (e.g., KernelSHAP~\cite{lundberg_unified_2017}) require evaluating a value function over many feature coalitions, which naively entails thousands of forward passes over highly overlapping sliding-window patches.

In this work, we propose an efficient perturbation-based explainability framework for \emph{patch-based 3D segmentation} by adapting KernelSHAP to volumetric CT data under two key principles: (1) \emph{localize} computation to a user-defined region of interest (ROI) and its receptive-field support, and (2) \emph{reuse} baseline computations whenever perturbations do not affect a given sliding-window patch. We further investigate how the definition of interpretable units (organs vs.\ geometric vs.\ organ-aware supervoxels) and the choice of scalar aggregation function (logit-weighted true/false positive scores, Dice, and Soft Dice) impact the qualitative appearance and the perturbation-based faithfulness of the resulting explanations.


%% file: sections/sec_related2.tex
A recent survey by Gipiškis et al.~\cite{gipiskis_explainable_2024} highlights two broad families that are most relevant to our setting: gradient-based methods and perturbation-/region-based methods. Compared to classification, segmentation explainability remains less explored and still lacks consolidated best practices, especially for \emph{(i)} defining clinically meaningful regions of analysis, \emph{(ii)} aggregating voxel-level evidence into robust scores, and \emph{(iii)} ensuring explanations remain interpretable in a structured anatomical setting.

\subsection{Gradient-based Attribution}
Gradient-based explanations extend saliency mechanisms to dense prediction. In segmentation, Grad-CAM-style methods have been adapted (e.g., Seg-Grad-CAM) to focus explanations on selected predicted regions rather than the full image~\cite{vinogradova_towards_2020}. Hasany et al.~\cite{hasany_guided_2024} show that the explanatory signal is highly layer-dependent (encoder bottleneck vs.\ decoder/output), motivating multi-layer inspection to capture how contextual anatomy shapes the final mask. More recently, FM-G-CAM~\cite{silva_fm-g-cam_2024} addresses the single-class limitation by integrating gradients from multiple classes, which is particularly relevant in multi-organ settings where inter-structure dependencies are clinically meaningful.

\subsection{Perturbation- and Region-based Methods}
Perturbation-based methods explain model behavior through systematic input modifications and their impact on the segmentation output. Early approaches include occlusion/sensitivity analysis to measure segmentation stability under localized masking or transformations~\cite{ankenbrand_sensitivity_2021}. In cardiac MRI, feature ablation studies further suggest that predictions for a target structure may depend on surrounding anatomical context~\cite{ayoob_peering_2025}, reinforcing the need for explanations that can disentangle target evidence from contextual cues.

However, classic local surrogate approaches such as LIME~\cite{ribeiro_why_2016} often struggle in segmentation due to instability and poor semantic alignment of perturbed regions. To mitigate this, Knab et al.~\cite{knab_beyond_2025} propose using anatomically meaningful superpixels derived from SAM to better match the perturbation units to realistic regions. SLICE~\cite{bora_slice_2024} targets variance reduction via stabilized sampling and superpixel selection, improving consistency across runs. Other works explicitly model context: Grid Saliency~\cite{hoyer_grid_2019} separates object evidence from contextual surroundings (at higher computational cost), while U-Noise-style approaches learn input-dependent noise masks and can be regularized for spatial smoothness~\cite{koker_u-noise_2021,okamoto_generating_2023}. Complementarily, MiSuRe~\cite{hasany_misure_2024} optimizes a minimally sufficient region that preserves segmentation quality (e.g., Dice) while shrinking the explanation, providing a fidelity-oriented notion of ``necessary'' evidence (in contrast to Grad-CAM/RISE~\cite{petsiuk_rise_2018}).

Overall, perturbation-based explanations are attractive because they are model-agnostic and can be aligned with clinically meaningful units, but their scalability and faithfulness remain open challenges. Chrabaszcz et al.~\cite{chrabaszcz_aggregated_2024} propose Agg\textsuperscript{2}Exp, aggregating voxel-level gradient attributions in 3D, and empirically highlight that gradient aggregation can provide more scalable and faithful insights than computationally heavy perturbation schemes in complex multi-class volumes.


%% file: sections/sec_methods2.tex
In this section, we present our explainability framework for interpreting predictions of patch-based 3D segmentation by adapting KernelSHAP~\cite{lundberg_unified_2017} to volumetric CT data.
Given the high dimensionality of voxel-level features, we estimate Shapley attributions over \emph{interpretable units} (supervoxels or organs) and restrict the analysis to a \emph{region of interest} (ROI).
Our pipeline is defined by the following components:
(i) ROI definition and corresponding receptive-field (RF) support for spatially localized computation;
(ii) a partition of the volume into $M$ interpretable units;
(iii) a perturbation operator that removes selected units by masking;
(iv) an ROI-restricted scalar \emph{value function} used by KernelSHAP to score coalitions;
(v) an efficient coalition-evaluation scheme for patch-based predictors based on sliding-window patch caching.

\subsection{ROI Definition and Receptive-Field Support}
\label{ssec:roi_rf_methods}

Our explanations are conditioned on a baseline segmentation that the user seeks to interpret.
In practice, an expert selects a subset of the model prediction for the target class (e.g., a connected component or a clinically relevant sub-region), thus defining an ROI mask $R(x)\in\{0,1\}$. Then, we compute the minimal axis-aligned \emph{cubic} bounding box $B$ enclosing all voxels such that $R(x)=1$.
Since inference is performed patch-wise (e.g., sliding-window evaluation), predictions inside $B$ depend only on a finite neighborhood of input voxels.
We therefore define an enlarged box $B_{\mathrm{RF}}$ by dilating $B$ to conservatively cover the input support that can influence predictions within $B$
(e.g., using the patch radius along each axis).
All perturbations and forward passes are restricted to $B_{\mathrm{RF}}$, while voxels outside are kept fixed.
This reduces computation without compromising faithfulness of attributions within the ROI.

\subsection{Interpretable Unit Definition}
\label{ssec:interpretable_units_methods}

KernelSHAP operates on $M$ binary features; therefore, the RF-supported crop $B_{\mathrm{RF}}$ is partitioned into interpretable units
$\{U_j\}_{j=1}^{M}$ with $U_j \subseteq B_{\mathrm{RF}}$.
These units are generated automatically inside $B_{\mathrm{RF}}$ by either algorithmic tessellation or by intersecting that tessellation with organ masks from TotalSegmentator; they are not manually drawn for each explanation.
We favor anatomy-aware variants because alignment with organ boundaries generally makes the resulting attributions easier to interpret clinically, even if no partition can perfectly match the model's latent internal representation.
We consider the following three partitions.

\smallskip\noindent
\textbf{Full Organs.}
Each anatomical structure segmented by TotalSegmentator~\cite{wasserthal_totalsegmentator_2022} is treated as one unit (background excluded).
This maximizes semantic interpretability and minimizes $M$, which typically improves KernelSHAP stability under a fixed sampling budget.

\begin{figure}
  \centering
  \includegraphics[width=0.3721\linewidth]{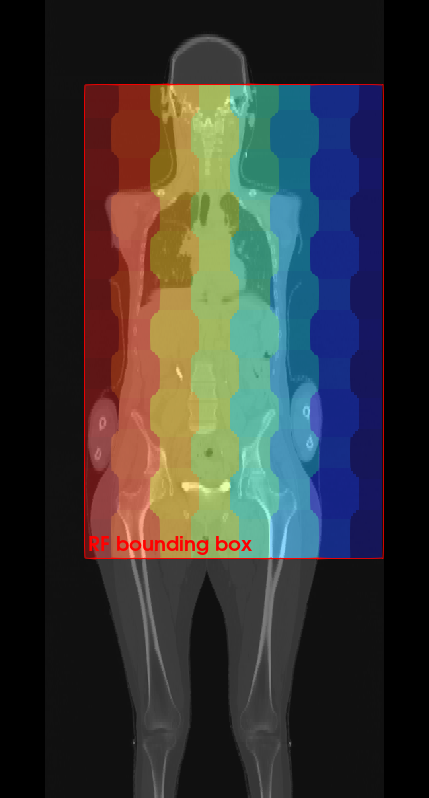}\hfill
  \includegraphics[width=0.4\linewidth]{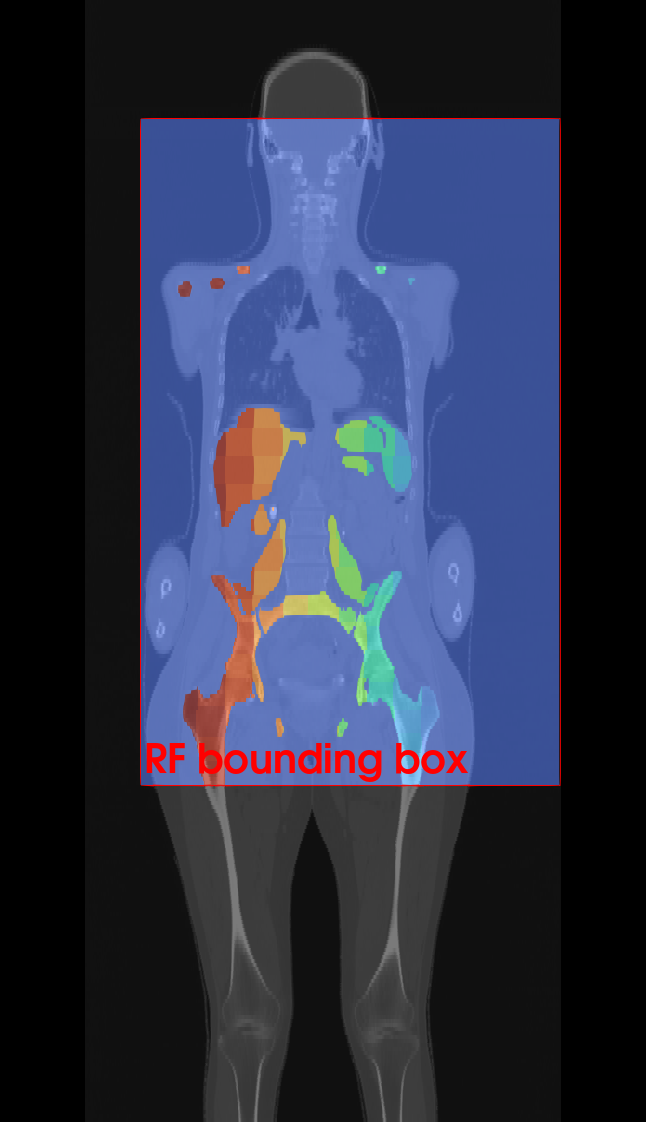}
  \caption{Coronal examples of (left) Regular FCC supervoxels and (right) Hybrid (organ-aware) FCC supervoxels within the RF-supported crop $B_{\mathrm{RF}}$.
  Regular FCC provides uniform geometry-driven units, whereas the Hybrid variant further splits FCC cells along organ boundaries to improve anatomical alignment.}
  \label{fig:sv_coronal}
\end{figure}

\smallskip\noindent
\textbf{Regular.}
We tessellate $B_{\mathrm{RF}}$ using a Face-Centred Cubic (FCC) lattice defined in physical coordinates (mm)~\cite{dardouillet_explainability_2022}.
Each voxel is assigned to its nearest lattice center, yielding approximately isotropic supervoxels in real-world units even under anisotropic voxel spacing.
A single scale parameter $S$ (mm) controls granularity. This organ-agnostic partition serves as a geometrically regular baseline (see an example in Fig.\ref{fig:sv_coronal}).

\smallskip\noindent
\textbf{Hybrid.}
\label{subsec:oa_fcc_methods}
To combine geometric regularity with anatomical constraints, we start from the FCC tessellation and subdivide each FCC cell according to organ labels
obtained from TotalSegmentator~\cite{wasserthal_totalsegmentator_2022}.
Voxels belonging to different organs within the same FCC cell are assigned distinct unit IDs (background excluded), preventing cross-structure mixing while preserving
physical-space isotropy (see an example in Fig.\ref{fig:sv_coronal}).
This design is conceptually related to organ-splitting strategies such as~\cite{chrabaszcz_aggregated_2024}, with the key difference that the underlying grid is FCC
rather than voxel-space cubes.

\subsection{Perturbation Operator}
\label{ssec:perturbation_methods}

Let $\mathbf{m}\in\{0,1\}^{M}$ denote a coalition mask, where $m_j=1$ keeps unit $U_j$ and $m_j=0$ removes it.
We implement hard masking in intensity space: for each removed unit, all voxels in $U_j$ are set to a masking value $b$. In CT we use $b=-1024$ HU, i.e., an air-equivalent value at the lower end of the Hounsfield scale that is commonly used for background/outside-body voxels. This gives a physically meaningful ``removal'' baseline without introducing spurious high-density tissue and is kept fixed throughout both attribution generation and evaluation.
Let $u(x)\in\{1,\dots,M\}$ denote the unit index of voxel $x\in B_{\mathrm{RF}}$. The perturbed input is
\begin{equation}
X^{(\mathbf{m})}(x) =
\begin{cases}
X(x), & m_{u(x)}=1,\\
b, & m_{u(x)}=0,
\end{cases}
\qquad x\in B_{\mathrm{RF}}.
\end{equation}
Voxels outside $B_{\mathrm{RF}}$ are never perturbed and are copied from the original input.
Although hard masking may introduce out-of-distribution artifacts, it provides an explicit ``removal'' intervention, making the resulting Shapley attributions
straightforward to interpret as contributions of the original signal within each unit.

\subsection{Aggregation Strategies for KernelSHAP Scoring}
\label{ssec:aggregation_methods}

KernelSHAP requires a scalar value function $v(\mathbf{m})$ for each coalition.
Let $f$ denote the segmentation network, producing voxel-wise logits $Z^{(\mathbf{m})}=f(X^{(\mathbf{m})})$.
For a target class $t$, define the perturbed hard prediction
\begin{equation}
P_{\mathbf{m}}(x) = \mathbbm{1}\!\left[\arg\max_{c} Z^{(\mathbf{m})}_{c}(x) = t\right],
\end{equation}
and the baseline (unperturbed) prediction $P_0$ as the special case $\mathbf{m}=\mathbf{1}$.
All scores are computed \emph{within the ROI} defined by $R(x)$; let $|R|=\sum_x R(x)$ and let $z^{(\mathbf{m})}_{t}(x)$ be the target-class logit. We compute the following scores to evaluate each coalition $\mathbf{m}$.

\smallskip\noindent
\textbf{True Positive Score.}
It rewards logit support for voxels correctly predicted as positive in the baseline segmentation within the ROI:
\begin{equation}
\label{eq:agg_tp_method}
S_{\mathrm{TP}}(\mathbf{m}) =
\frac{1}{|R|}\sum_{x} R(x)\,\mathbbm{1}\!\left[P_{\mathbf{m}}(x)=1 \wedge P_{0}(x)=1\right]\, z^{(\mathbf{m})}_{t}(x).
\end{equation}

\smallskip\noindent
\textbf{False Positive Score.}
It penalizes logit support for voxels incorrectly predicted as positive in the baseline segmentation within the ROI:
\begin{equation}
\label{eq:agg_fp_method}
S_{\mathrm{FP}}(\mathbf{m}) =
-\frac{1}{|R|}\sum_{x} R(x)\,\mathbbm{1}\!\left[P_{\mathbf{m}}(x)=1 \wedge P_{0}(x)=0\right]\, z^{(\mathbf{m})}_{t}(x).
\end{equation}

\smallskip\noindent
\textbf{Dice Score.}
It exploits Dice similarity to quantify the agreement between perturbed and baseline predictions:
\begin{equation}
\label{eq:agg_dice_method}
S_{\mathrm{Dice}}(\mathbf{m}) =
\frac{2\,\bigl|\,(P_{\mathbf{m}}\!\odot R)\cap(P_{0}\!\odot R)\,\bigr|}
{\|P_{\mathbf{m}}\!\odot R\|_{1}+\|P_{0}\!\odot R\|_{1}+\varepsilon},
\end{equation}
where $\odot$ denotes element-wise masking by the ROI and $\varepsilon>0$ avoids division by zero.

\smallskip\noindent
\textbf{Soft Dice Score.}
It rewards logit support on voxels consistent with the baseline segmentation and penalizes newly introduced positives within the ROI:
\begin{equation}
\label{eq:agg_softdice_method}
S_{\mathrm{Soft}}(\mathbf{m}) =
\frac{1}{|R|}\sum_{x} R(x)\,\mathbbm{1}\!\left[P_{\mathbf{m}}(x)=1\right]\, w(x)\, z^{(\mathbf{m})}_{t}(x).
\end{equation}
where $w(x)=2P_0(x)-1$, i.e., +1 if voxel $x$ is in the baseline segmentation and -1 otherwise.
In other words, $w(x)$ acts as a signed agreement label: logits on voxels already present in the baseline mask are rewarded, whereas logits on newly activated voxels are penalized, making $S_{\mathrm{Soft}}$ a soft surrogate of Dice agreement with the baseline segmentation.


\subsection{Efficient Coalition Evaluation via Sliding-Window Patch Caching}
\label{ssec:patch_caching_methods}

KernelSHAP requires evaluating the value function over many coalitions, hence many forward passes of the segmentation model on perturbed inputs.
For patch-based predictors using sliding-window inference (e.g., \texttt{nnU-Net}), a naive implementation would recompute logits for every overlapping patch at every sampled coalition, which is typically prohibitive even when restricting computation to $B_{\mathrm{RF}}$.
To reduce redundant computation, we cache baseline patch logits and reuse them whenever a coalition does not affect a given patch.

\smallskip\noindent
\textbf{Baseline cache construction.}
We first run a single sliding-window inference on the unperturbed input restricted to $B_{\mathrm{RF}}$.
Let $\mathcal{S}$ denote the set of sliding-window patch extractors (slices) used by the inference routine.
For each patch location $s\in\mathcal{S}$, we store the predicted patch logits in a dictionary keyed by the patch spatial coordinates (slice key).
To limit GPU memory usage, cached logits are stored in CPU memory.

\smallskip\noindent
\textbf{Cached inference under perturbations.}
For a coalition $\mathbf{m}$, we define a binary perturbation mask $\mathbf{M}^{(\mathbf{m})}$ over $B_{\mathrm{RF}}$ such that $\mathbf{M}^{(\mathbf{m})}(x)=1$ if voxel $x$ is masked to the masking value $b=-1024$ (see Section~\ref{ssec:perturbation_methods}).
During sliding-window fusion, for each patch slice $s\in\mathcal{S}$ we check whether the coalition affects that patch region.
If $\sum_{x\in s}\mathbf{M}^{(\mathbf{m})}(x)=0$ (no perturbed voxels in the patch), we retrieve the cached baseline logits (cache hit); otherwise, we recompute logits by forwarding the perturbed patch only (cache miss).
Cached and recomputed patch logits are fused exactly as in standard \texttt{nnU-Net} inference (Gaussian-weighted accumulation followed by normalization), ensuring that caching does not alter the aggregation scheme.

\smallskip\noindent
\textbf{Expected savings and trade-offs.}
Let $h$ be the cache hit rate, i.e., the fraction of patches retrieved from the cache for a given coalition.
Under an idealized constant per-patch cost model and ignoring overheads, the forward-pass time scales with $(1-h)$, yielding an approximate speedup of $1/(1-h)$ relative to recomputing all patches.
In practice, the realized gain is lower due to cache lookups, intersection tests, and CPU--GPU transfers, but caching remains most effective when perturbations are spatially localized (as typically occurs when interpretable units correspond to anatomically constrained supervoxels within a focused ROI).
The main trade-off is memory, since the cache stores logits for all sliding-window patches covering $B_{\mathrm{RF}}$.

%% file: sections/sec_design2.tex
This section details the experimental setup used to compute KernelSHAP attribution maps for the nnU-Net segmentation model and to quantitatively evaluate their faithfulness. 

\subsection{Data and Segmentation Task}
\label{subsec:data_task_methods}

As testbed for our explainability framework, we use a clinical dataset of whole-body CT images for segmentation of lymph nodes and spleen in the context of Total Marrow and Lymphoid Irradiation (TMLI) treatment planning. The full dataset comprises \textbf{40 patients} affected by hematological malignancies and treated with non-myeloablative TMLI.
CT scans were acquired in free-breathing and without contrast using a clinical scanner (slice thickness of 5~mm). Volumes have an average shape of $237 \times 512 \times 512$ and anisotropic spacing of approximately $5.0~\mathrm{mm} \times 1.17~\mathrm{mm} \times 1.17~\mathrm{mm}$.

The \textbf{segmentation target} is defined as the union of \textbf{lymph nodes (CTV\_LN)} and \textbf{spleen (CTV\_Spleen)}, which are challenging due to high inter-patient variability and complex geometries. 

The dataset is split into \textbf{32 training} and \textbf{8 test} volumes. We train a 3D nnU-Net~\cite{isensee_nnu-net_2021} segmentation model on the training set, following the setup described in~\cite{brioso2025investigating}. 

On the 8 test volumes, we evaluated the proposed explainability framework using three different interpretable-unit definitions, i.e., Full Organs, Regular, and Hybrid (see Section~\ref{ssec:interpretable_units_methods}) and four aggregation/score functions, i.e., True Positive, False Positive, Dice, Soft Dice (see Section~\ref{ssec:aggregation_methods}). 
During attribution computation, nnU-Net test-time augmentation was disabled to ensure deterministic coalition evaluations, and inference was accelerated via patch caching (Section~\ref{ssec:patch_caching_methods}).


\subsection{KernelSHAP Convergence Validation and Sampling Budget}
\label{subsec:convergence_validation_methods}

To ensure reliable Shapley attributions, we empirically validated the stability of the KernelSHAP approximation as a function of the sampling budget $n$ (number of coalitions). KernelSHAP fits a weighted linear surrogate model to coalition evaluations of the nnU-Net value function. We increased $n$ and monitored surrogate stability using: (i) \textit{coefficient stability} (relative $\ell_1$ change of attribution vectors between successive budgets), (ii) \textit{local accuracy} (residual between the coalition value and the sum of attributions), (iii) \textit{numerical stability} (condition number of the weighted design matrix), and (iv) \textit{generalization of the surrogate} on held-out coalitions (MAE and $R^2$).

Based on this stability analysis, we selected a conservative budget of $\mathbf{n = 2000}$ coalitions for the \textit{Regular} and \textit{Hybrid} settings (typically involving several hundred features), while $\mathbf{n = 1000}$ was sufficient for \textit{Full Organs} due to the much smaller feature space ($M \approx 9$).

\subsection{Attribution Map Evaluation Metrics}
\label{subsec:evaluation_metrics_methods}

We evaluate attribution maps primarily through \textbf{perturbation-curve faithfulness}, balancing computational feasibility in 3D with interpretability and direct alignment to the model behavior.

\smallskip\noindent
\textbf{MoRF/LeRF perturbation protocol.}
For a given volume and configuration, interpretable units are ranked by their SHAP values. Starting from the unperturbed input, we iteratively remove (mask) units according to two complementary orderings: \textit{Most-Relevant-First} (MoRF, descending SHAP) and \textit{Least-Relevant-First} (LeRF, ascending SHAP). After each step, we recompute the model output and the corresponding scalar score using the same ROI-restricted aggregation/value function used for KernelSHAP (Section~\ref{ssec:aggregation_methods}) and the same masking baseline ($b=-1024$ HU; Section~\ref{ssec:perturbation_methods}). This yields two curves, $s_{\mathrm{MoRF}}(k)$ and $s_{\mathrm{LeRF}}(k)$, as a function of the amount of removed evidence. Therefore, the reported perturbation-curve metrics validate attribution faithfulness under the exact perturbation operator used to generate the explanations.

\smallskip\noindent
\textbf{Area Over the Perturbation Curve (AOPC).}
From the MoRF curve, we compute AOPC as the average degradation in score caused by removing highly-ranked units~\cite{samek_evaluating_2015}:
\begin{equation}
\mathrm{AOPC} = \frac{1}{K}\sum_{k=1}^{K}\big(s(0) - s_{\mathrm{MoRF}}(k)\big),
\end{equation}
where $s(0)$ is the unperturbed score and $K$ is the number of perturbation steps. Higher AOPC indicates that the attribution map successfully identifies units whose removal most strongly impacts the model score.

\smallskip\noindent
\textbf{Area Between Perturbation Curves (ABPC).}
To quantify how well an attribution method separates important from unimportant evidence, we compute ABPC as the average gap between the LeRF and MoRF trajectories:
\begin{equation}
\mathrm{ABPC} = \frac{1}{K}\sum_{k=1}^{K}\big(s_{\mathrm{LeRF}}(k) - s_{\mathrm{MoRF}}(k)\big).
\end{equation}
Higher ABPC indicates stronger discrimination: removing low-ranked units leaves the score comparatively intact, while removing high-ranked units rapidly degrades it.

Because different aggregation functions (and different volumes) can induce different score ranges, we also report normalized variants of AOPC/ABPC. For each curve, we rescale scores to $[0,1]$ using the attainable range induced by the perturbation endpoints (analogous in spirit to normalized faithfulness metrics such as NAOPC~\cite{edin_normalized_2025}):
\begin{equation}
\tilde{s}(k) = \frac{s(k) - s_{\min}}{s_{\max} - s_{\min} + \epsilon},
\end{equation}
with $s_{\max}=s(0)$ and $s_{\min}$ given by the fully-perturbed score (all units removed), computed separately for each case, configuration, and aggregation function.

Finally, for interpretability across heterogeneous tessellations, perturbation curves are plotted against the fraction (or number) of interpretability units removed.

%% file: sections/sec_results2.tex
This section reports the results obtained with the KernelSHAP attribution framework described in Section~\ref{sec:methods}. We first provide a qualitative inspection of attribution maps across different interpretable-unit definitions and aggregation functions. We then summarize the quantitative evaluation based on MoRF/LeRF perturbation-curve metrics. Finally, we report the computational gains obtained through patch caching. All results are computed on the eight validation volumes defined in the experimental design.

\subsection{Qualitative Analysis of Attribution Maps}
\label{sec:qualitative_results_xai}

\begin{figure}
    \centering
    \includegraphics[width=90mm]{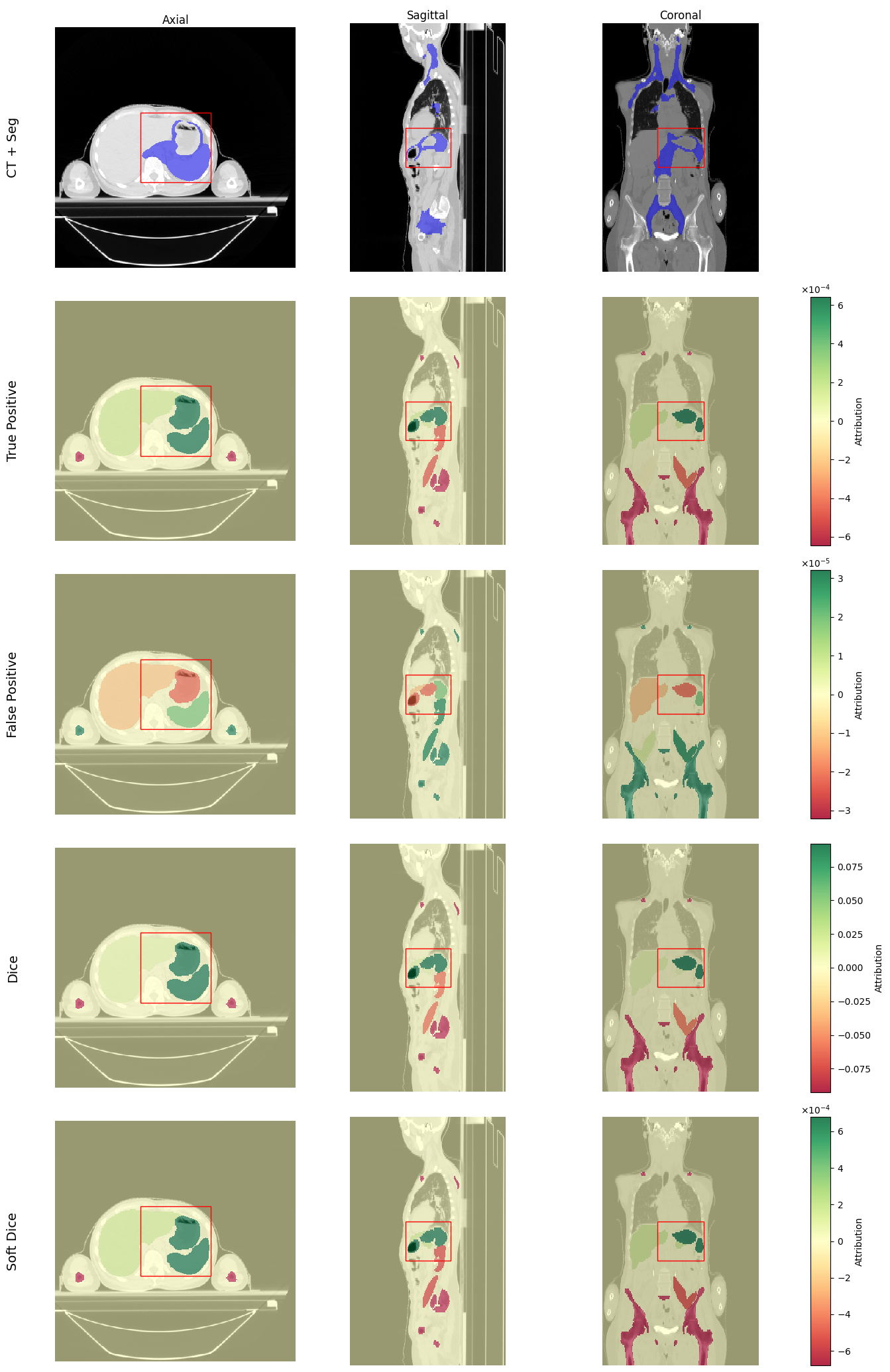}
    \caption{Qualitative attribution maps for \textbf{Full Organs} (volume 7), comparing aggregation functions within the ROI.}
    \label{fig:qualitative_full_organs_xai}
\end{figure}

\begin{figure}
    \centering
    \includegraphics[width=90mm]{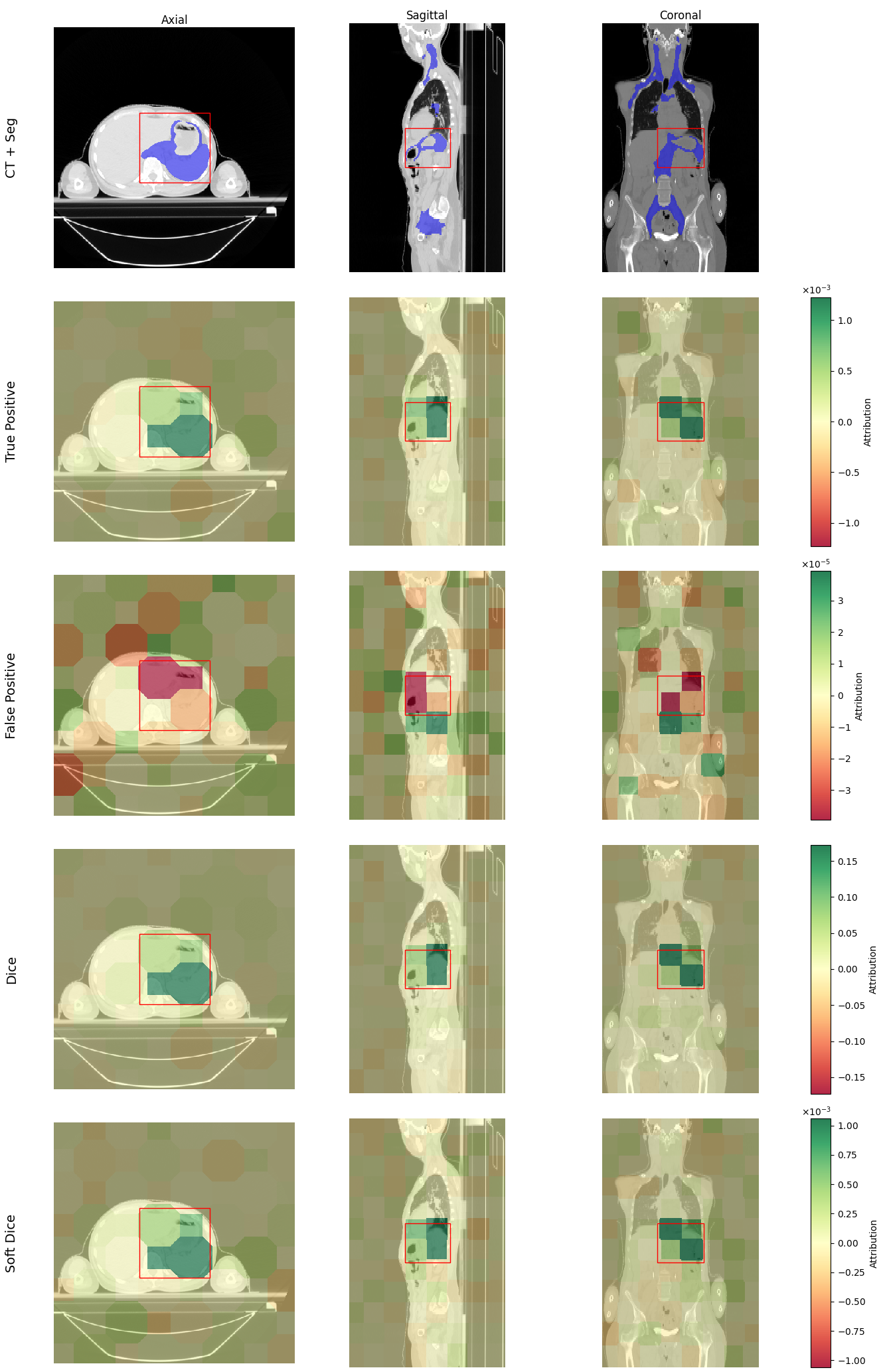}
    \caption{Qualitative attribution maps for \textbf{Regular (FCC)} supervoxels (volume 7).}
    \label{fig:qualitative_regular_xai}
\end{figure}

\begin{figure}
    \centering
    \includegraphics[width=90mm]{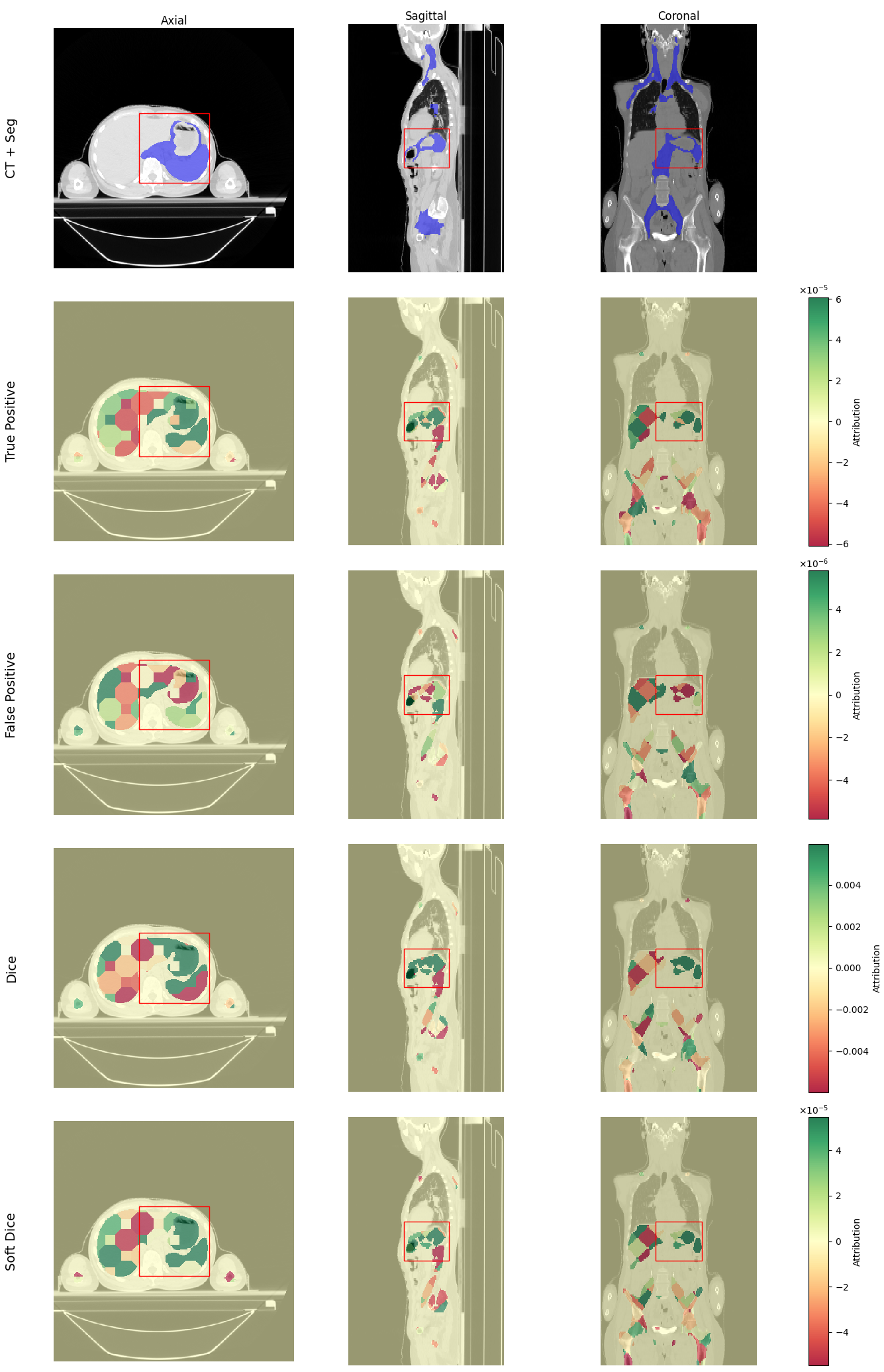}
    \caption{Qualitative attribution maps for \textbf{Hybrid (Organ-Aware FCC)} supervoxels (volume 7).}
    \label{fig:qualitative_hybrid_xai}
\end{figure}

Visual inspection of the attribution maps highlights how the choice of interpretable units (supervoxels) and the value-function aggregation affect the resulting explanations. Figures~\ref{fig:qualitative_full_organs_xai}--\ref{fig:qualitative_hybrid_xai} show representative examples for \textbf{volume 7}. Attribution magnitudes are generally small due to ROI restriction and score normalization. Both strongly positive values (units supporting the baseline prediction) and strongly negative values (units opposing it, e.g., by reducing Dice or increasing false positives) indicate high relevance.

\smallskip\noindent
\textbf{Full Organs (Fig.~\ref{fig:qualitative_full_organs_xai}).}
Attributions are constrained by organ boundaries. The TP-, Dice-, and Soft-Dice-based maps are visually consistent and emphasize regions near the ROI that support the baseline segmentation. In contrast, the FP-based map shows an approximately inverted pattern, highlighting regions whose presence contributes to spurious activations. This suggests that anatomically coherent regions may play a dual role, stabilizing correct predictions while also inducing false positives depending on local context.

\smallskip\noindent
\textbf{Regular (FCC) supervoxels (Fig.~\ref{fig:qualitative_regular_xai}).}
Due to the large and spatially uniform units, maps exhibit stronger local effects and sharper transitions. TP, Dice, and Soft Dice again emphasize evidence within or close to the ROI supporting the baseline output. The FP map reveals negative contributions concentrated within the ROI (spurious activations) and occasional positive contributions near boundaries, consistent with a mix of destabilizing and stabilizing effects. Overall, explanations appear noisier, which is expected from an organ-agnostic tessellation and the coarser effective resolution.

\smallskip\noindent
\textbf{Hybrid (Organ-Aware FCC) supervoxels (Fig.~\ref{fig:qualitative_hybrid_xai}).}
Hybrid units preserve organ semantics while enabling finer within-organ granularity. Relative to Full Organs, TP/Dice/Soft-Dice maps capture more detailed intra-organ variations. Importantly, FP-based attributions show marked heterogeneity inside single organs, where different subregions may contribute with opposite signs. This behavior is consistent with localized mechanisms leading to spurious predictions and may provide a more actionable explanation for debugging false positives.

\subsection{Quantitative Evaluation using Perturbation Curves}
\label{sec:quantitative_results_xai}

\begin{figure}
    \centering
    \includegraphics[width=0.95\columnwidth]{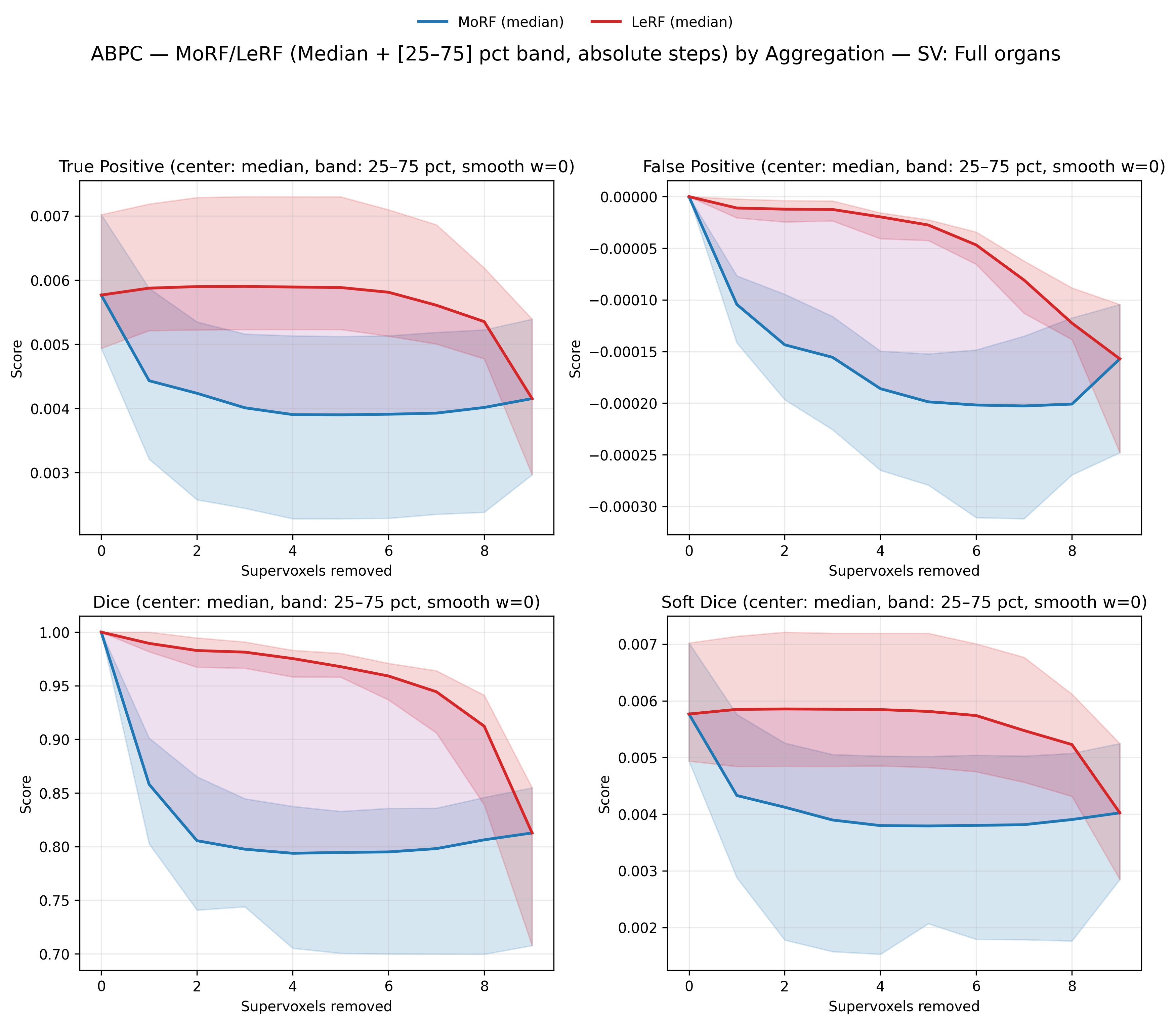}
    \caption{MoRF and LeRF curves (median $\pm$ IQR) for \textbf{Full Organs}.}
    \label{fig:abpc_full_xai}
\end{figure}

\begin{figure}
    \centering
    \includegraphics[width=0.95\columnwidth]{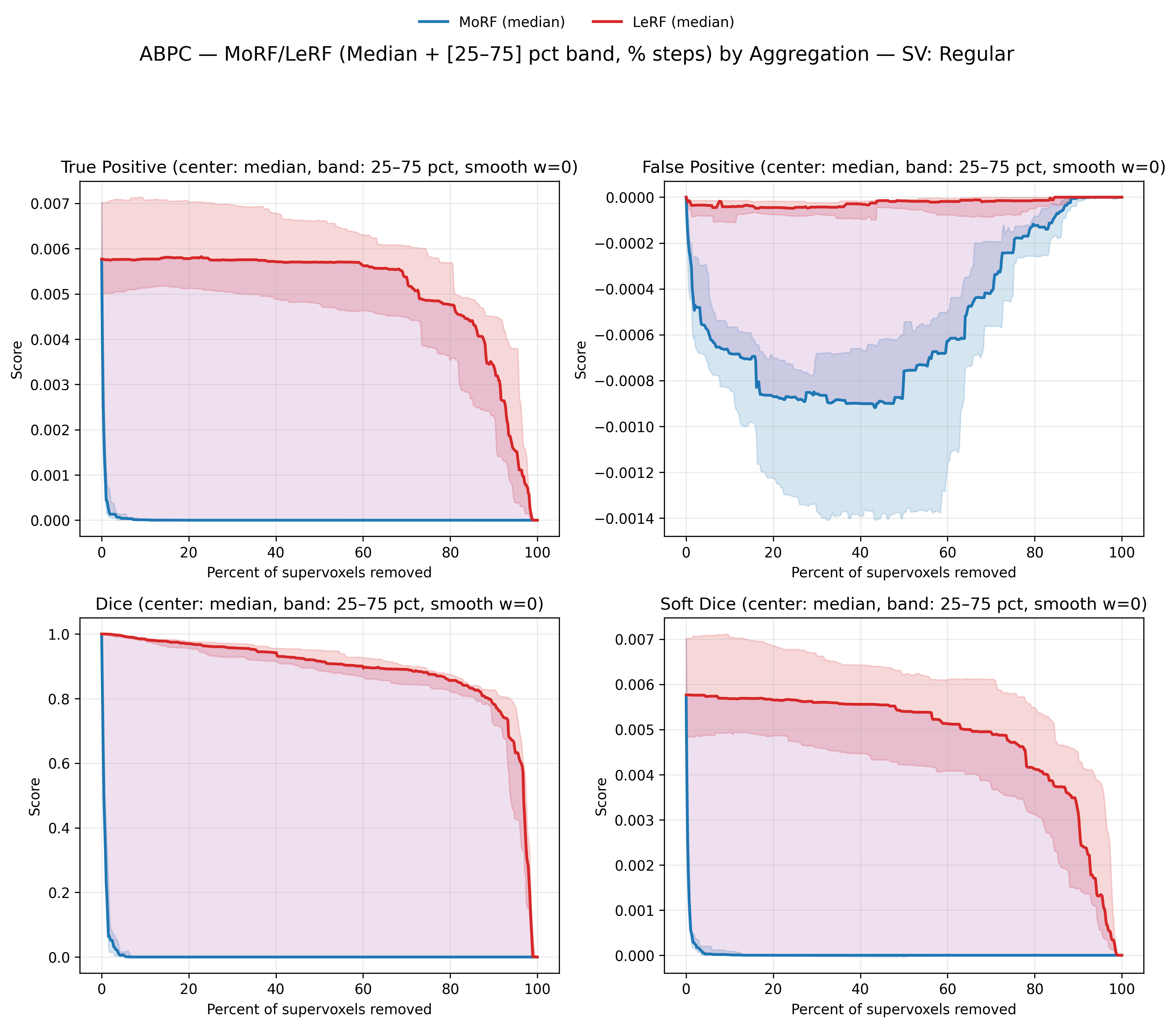}
    \caption{MoRF and LeRF curves (median $\pm$ IQR) for \textbf{Regular (FCC)} supervoxels.}
    \label{fig:abpc_regular_xai}
\end{figure}

\begin{figure}
    \centering
    \includegraphics[width=0.95\columnwidth]{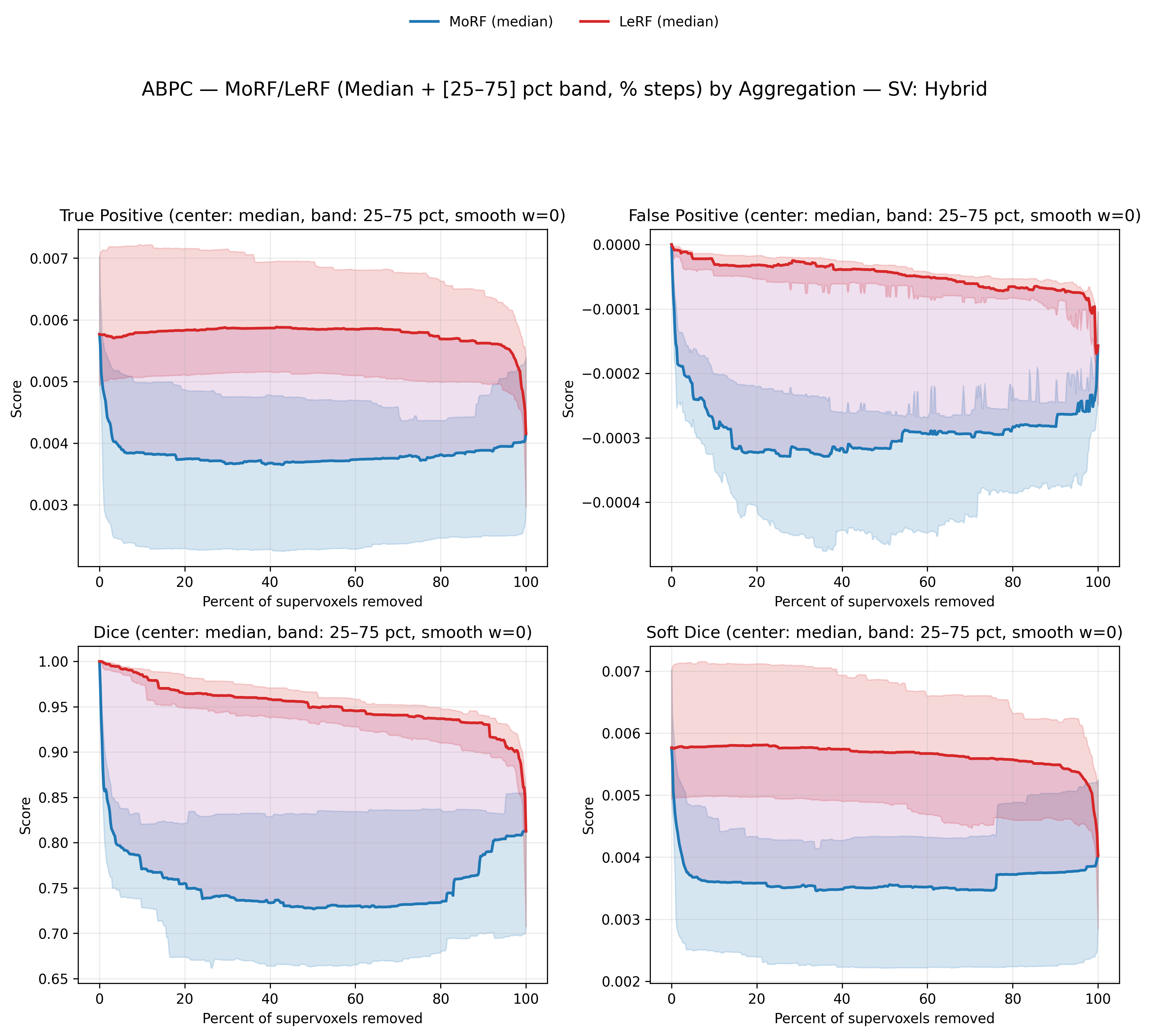}
    \caption{MoRF and LeRF curves (median $\pm$ IQR) for \textbf{Hybrid} supervoxels.}
    \label{fig:abpc_hybrid_xai}
\end{figure}

Figures~\ref{fig:abpc_full_xai}--\ref{fig:abpc_hybrid_xai} report MoRF and LeRF perturbation curves (median $\pm$ IQR), while Table~\ref{tab:abpc_aopc_results_xai} summarizes ABPC/AOPC and their normalized variants averaged over the eight validation cases.

Across TP, Dice, and Soft Dice aggregations, \textbf{Regular} supervoxels achieve the highest raw and normalized ABPC/AOPC values. These results are mainly due to two factors. First, the FCC tessellation is organ-agnostic and typically spans a \emph{larger overall spatial support} than organ-constrained partitions, so that successive perturbations remove (or affect) a broader portion of the input volume. Second—and most critically—FCC units may \emph{overlap the segmentation target} (i.e., voxels that are maximally correlated with the value function). Perturbing units that contain the target can induce a direct and substantial drop in TP/Dice/Soft Dice, which inflates MoRF--LeRF separability and, consequently, ABPC/AOPC (including their normalized variants). Therefore, cross-configuration comparisons based solely on perturbation metrics are not entirely fair, as they conflate attribution quality with feature definition and target overlap.

From an interpretability standpoint, this also highlights a practical limitation of Regular: attributing importance to units that effectively "contain the answer" (the target) tends to yield explanations that are less clinically actionable, because they emphasize the model's sensitivity to removing the target itself rather than revealing anatomically meaningful contextual drivers.

For the \textbf{False Positive} aggregation, \textbf{Hybrid} supervoxels obtain the best performance under normalized metrics (nABPC and nAOPC). This suggests that, at comparable granularity, combining anatomical constraints with within-organ partitioning improves the ability to localize subregions specifically responsible for spurious activations, compared to purely geometric FCC tessellations.

Finally, TP and Soft Dice show consistent trends across supervoxel types, whereas Dice (computed on binarized masks) tends to accentuate differences between configurations, likely due to thresholding effects and discrete changes in overlap.

\subsection{Computational Performance with Caching}
\label{sec:caching_results_xai}

Patch caching (Section~\ref{ssec:patch_caching_methods}) substantially reduces redundant computation during coalition evaluation. Averaged over the eight validation cases, \textbf{Full Organs} achieves an average cache hit ratio of $32.4\% \pm 6.3\%$ with an inference time of $3.58\text{s} \pm 0.47\text{s}$ per sample, yielding a total runtime of \textbf{1h 01m 45s} for $n=1000$ coalitions. \textbf{Regular (FCC)} shows lower cache reuse ($15.0\% \pm 1.4\%$) and higher inference time ($4.41\text{s} \pm 0.57\text{s}$), resulting in the longest total runtime (\textbf{2h 31m 47s}) for $n=2000$. \textbf{Hybrid} preserves high cache reuse ($30.2\% \pm 5.2\%$) with inference time comparable to Full Organs ($3.57\text{s} \pm 0.43\text{s}$), leading to a total runtime of \textbf{2h 02m 36s} for $n=2000$.
These results confirm that caching is most effective when perturbations remain spatially localized relative to the sliding-window patch grid, as in organ-constrained or organ-aware unit definitions. Overall, \textbf{Hybrid} provides a favorable trade-off between explanation granularity (larger feature space requiring higher sampling budgets) and computational efficiency (high cache reuse).

\begin{table}[t]
\centering
\caption{Summary of perturbation-curve metrics (mean over 8 validation cases). Best results per row are in bold.}
\label{tab:abpc_aopc_results_xai}

\renewcommand{\arraystretch}{1.05}
\setlength{\tabcolsep}{5pt}

\begin{tabular}{@{} c l c c c @{}}
\toprule
\textbf{Metric} & \textbf{Aggregation} & \textbf{Full organs} & \textbf{Regular} & \textbf{Hybrid} \\
\midrule
\multirow{4}{*}{\textbf{ABPC}}
 & True Positive  & 1.576e-03 & \textbf{4.980e-03} & 2.102e-03 \\
 & False Positive & 4.065e-04 & \textbf{8.095e-04} & 2.603e-04 \\
 & Dice           & 1.535e-01 & \textbf{8.387e-01} & 2.022e-01 \\
 & Soft Dice      & 1.524e-03 & \textbf{4.753e-03} & 2.005e-03 \\
\midrule
\multirow{4}{*}{\textbf{nABPC}}
 & True Positive  & 0.6776 & \textbf{0.8341} & 0.8112 \\
 & False Positive & 0.5259 & 0.4827 & \textbf{0.6653} \\
 & Dice           & 0.5869 & \textbf{0.8388} & 0.6960 \\
 & Soft Dice      & 0.6227 & \textbf{0.7787} & 0.7574 \\
\midrule
\multirow{4}{*}{\textbf{AOPC}}
 & True Positive  & 1.565e-03 & \textbf{5.751e-03} & 2.158e-03 \\
 & False Positive & 4.504e-04 & \textbf{8.593e-04} & 3.118e-04 \\
 & Dice           & 2.017e-01 & \textbf{9.857e-01} & 2.581e-01 \\
 & Soft Dice      & 1.737e-03 & \textbf{5.811e-03} & 2.291e-03 \\
\midrule
\multirow{4}{*}{\textbf{nAOPC}}
 & True Positive  & 0.7625 & \textbf{0.9898} & 0.8948 \\
 & False Positive & 0.7091 & 0.5251 & \textbf{0.8165} \\
 & Dice           & 0.7745 & \textbf{0.9858} & 0.9002 \\
 & Soft Dice      & 0.7537 & \textbf{0.9658} & 0.9002 \\
\bottomrule
\end{tabular}
\end{table}

%% file: sections/sec_conclusions2.tex
We presented an efficient perturbation-based explainability framework for patch-based 3D medical image segmentation by adapting KernelSHAP to volumetric CT. Our pipeline localizes coalition evaluations to a user-defined ROI and its receptive-field support, and accelerates nnU-Net sliding-window inference through patch logit caching that reuses baseline predictions whenever a coalition does not affect a patch. This substantially reduces redundant computation and makes Shapley-style attributions more practical in 3D settings.

Our results show that explanation meaning and quality are tightly coupled to the definition of \emph{interpretable units} and the \emph{aggregation/value function}. Perturbation-curve metrics (AOPC, ABPC, including normalized variants) highlighted clear trade-offs: Regular supervoxels often achieved the highest faithfulness due to their large spatial influence and frequent overlap with the target, but lacked anatomical alignment and may be less actionable clinically. Full Organs provided the most interpretable, stable high-level explanations with a minimal feature set, at the cost of limited granularity. Hybrid organ-aware supervoxels offered a compelling balance, preserving organ semantics while enabling finer within-organ resolution, and were particularly effective at exposing features associated with false positives under normalized metrics. In addition, the aggregation choice modulated the explanatory focus, emphasizing either stabilizing evidence (TP/Dice/Soft Dice) or destabilizing effects linked to spurious activations (FP), underscoring that different clinical questions require different objectives.

Several limitations remain. Validation was performed on only eight held-out volumes due to KernelSHAP’s computational cost, so the reported trends should be interpreted as preliminary and confirmed on larger, more diverse cohorts. The approach also inherits KernelSHAP assumptions, including a linear surrogate and sensitivity to the perturbation strategy; hard masking (zero-out to $-1024$ HU) may introduce out-of-distribution artifacts. Moreover, we did not include clinician-centered evaluation, so quantitative faithfulness/stability metrics do not directly translate to clinical usefulness. Finally, fixed supervoxel strategies may bias explanations when units misalign with clinically meaningful boundaries.

Future work will investigate data-driven, boundary-adherent 3D supervoxels (e.g., 3D SLIC \cite{achanta_slic_2012} or SEEDS \cite{bergh_seeds_2013,zhao_extending_2025}) to improve anatomical adherence while characterizing cost--regularity trade-offs. On the efficiency side, more sophisticated forms of caching (e.g., caching frequently occurring coalitions in low-overlap patches, common with Full Organs) could further reduce runtime. We also plan to explore more realistic in-distribution perturbations, such as inpainting-based removal.